\documentclass{article}

    \usepackage[final,nonatbib]{neurips_workshop_2021}

\usepackage[utf8]{inputenc} 
\usepackage[T1]{fontenc}    
\usepackage{hyperref}       
\usepackage{url}            
\usepackage{booktabs}       
\usepackage{amsfonts}       
\usepackage{nicefrac}       
\usepackage{microtype}      
\usepackage{xcolor}         
\usepackage{graphicx}
\usepackage{amsmath}
\usepackage{multirow}
\usepackage{subcaption}
\usepackage{authblk}

\title{AVoE: A Synthetic 3D Dataset on Understanding Violation of Expectation for Artificial Cognition}

\author[1]{\textbf{Arijit Dasgupta}}
\author[2]{\textbf{Jiafei Duan}}
\author[1]{\textbf{Marcelo H. Ang Jr}}
\author[2]{\textbf{Cheston Tan}}
\affil[1]{Department of Mechanical Engineering, National University of Singapore}
\affil[2]{Institute for Infocomm Research, A*STAR}

\begin{document}

\maketitle

\begin{abstract}
Recent work in cognitive reasoning and computer vision has engendered an increasing popularity for the Violation-of-Expectation (VoE) paradigm in synthetic datasets. Inspired by work in infant psychology, researchers have started evaluating a model's ability to discriminate between expected and surprising scenes as a sign of its reasoning ability. Existing VoE-based 3D datasets in physical reasoning only provide vision data. However, current cognitive models of physical reasoning by psychologists reveal infants create high-level abstract representations of objects and interactions. Capitalizing on this knowledge, we propose AVoE: a synthetic 3D VoE-based dataset that presents stimuli from multiple novel sub-categories for five event categories of physical reasoning. Compared to existing work, AVoE is armed with ground-truth labels of abstract features and rules augmented to vision data, paving the way for high-level symbolic predictions in physical reasoning tasks. \href{https://github.com/jiafei1224/AVoE}{\color{blue} AVoE: https://github.com/jiafei1224/AVoE}
\end{abstract}

\section{Introduction}

Physical reasoning systems built on the foundations of intuitive physics \& psychology are pivotal to creating machines that learn and think like humans \cite{lake2017building, adams2012mapping}. The ability to physically reason like humans forms a crucial gateway to multiple real-world applications ranging from robotic assistants, autonomous vehicles, and safe AI tools. These systems are guided by facets of core knowledge \cite{spelke2007core}, from object representation to numerical, spatial \& social reasoning. Infant psychologists theorised that human newborns have an in-built physical reasoning \cite{baillargeon2011infants, hespos2008young} and object representation \cite{kahneman1992reviewing,gordon1996s, stavans_lin_wu_baillargeon_2019} system guiding their learning of core knowledge. This has inspired researchers to approach the creation of physical reasoning systems as a start-up software embedded with core knowledge principles \cite{ullman2020bayesian}.

Predicting the future of a physical interaction given a scene prior has been the most common task in designing computational physical reasoning systems. One such task is the tower of falling objects, which has been as a test-bed for evaluating intuitive physics engines \cite{battaglia2013simulation, zhang2016comparative, lerer2016learning}. There have also been recent advancements in creating benchmarked datasets and models that combine multiple physical prediction tasks in a 3D environment \cite{duan2021space, bear2021physion, duan2021pip}. Physical reasoning has also been explored in Embodied AI \cite{duan2021survey} \& Visual Question Answering (VQA) \cite{wu2017visual} with public datasets like \textbf{CLEVRER} \cite{yi2019clevrer}, \textbf{CRAFT} \cite{ates2020craft} \& \textbf{TIWIQ} \cite{wagner2018answering} providing scenes of general and random interactions between objects and multiple questions concerning the physical outcome.

A parallel track complementary to future prediction tasks is the design of artificial agents that can measure the plausibility of physical scenes. An agent capable of physical reasoning should not only be able to predict the future but also recognise if a scene is \textit{possible} or \textit{impossible}. The Violation-of-Expectation (VoE) paradigm is an empirical diagnostic tool first implemented in developmental psychology studies \cite{baillargeon1985object, baillargeon1987object} to measure the surprise of infants when shown \textit{possible} or \textit{impossible} scenes. The studies found that infants as young as 4 months could express surprise at a constructed scene that violated the principle of object permanence: the tendency for an object to continually exist without dropping in and out of existence. This was akin to a magic show, and the surprise was measured by the looking time of the infants. VoE has since been used in an array of infant psychology experiments on a range of physical reasoning event categories \cite{baillargeon1990young, spelke1995spatiotemporal, dan2000development, wang2005detecting, kotovsky1994calibration}.

The work in VoE has encouraged recent computational development of models and datasets \cite{piloto2018probing, riochet2018intphys, smith2019modeling} that challenge artificial agents to differentiate between \textit{possible} and \textit{impossible} scenes in physical events. Instead of training with \textit{possible} and \textit{impossible} scenes, only \textit{possible} scenes \& interactions are shown in the training dataset consistently among the studies. This inherently replicates the experience of a normal growing infant who is usually exposed to \textit{possible} scenes. Therefore, the added challenge of an imbalanced and realistic training dataset means that researchers have to look beyond general end-to-end deep learning models and focus on generative and probabilistic approaches leveraging on inductive biases. VoE is not limited to physical reasoning, for it can be used in any reasoning task that has a clear binary distinction between \textit{surprising} and \textit{expected}. \textbf{AGENT} \cite{shu2021agent} \& \textbf{BIB} \cite{gandhi2021bib} are two similar and recently developed datasets that train and evaluate an agent's ability to recognise another agent's goal preference \cite{woodward1998infants}, action efficiency \cite{gergely1995taking} and more. Taking inspiration from the increasing use of VoE for computational reasoning tasks, we propose AVoE: a novel VoE-based 3D dataset of five different event categories of physical reasoning. Our contributions for AVoE are two-fold:

\begin{itemize}
    \item Existing VoE-based physical reasoning datasets \cite{piloto2018probing, riochet2018intphys, smith2019modeling} only provide vision-based data. While vision still forms the main sensing modality of AVoE, we take a step further by providing additional scene-wise ground-truth metadata of abstract features and rules which models can exploit for higher-level symbolic prediction of the physical reasoning tasks.
    \item Compared to the existing datasets, AVoE contains scene stimuli in multiple novel sub-categories for each event category, adding to its diversity and complexity.
\end{itemize}

\section{Related Work}

The intersection of computer vision and physical reasoning is heavily grounded in the literature of psychology. The age in which infants adopt core principles of persistence, inertia \& gravity \cite{lin2020infants} have been widely studied in a variety of event categories. For example, the \textbf{barrier} event is an event category illustrating (or violating) the constraint of solidity. In studies that implemented the barrier event to infants \cite{baillargeon1990young, spelke1992origins}, psychologists placed a solid barrier with an object on one side and the \textit{surprising} event occurred when the infant was made to believe that the object could pass through the barrier. Like the barrier event, there are other events like \textbf{containment} \cite{hespos2001infants, wang2005detecting, mou2017container} where a \textit{surprising} scene was demonstrated with a taller object being fully contained in a shorter container, violating solidity constraints. Continuity constraints were also violated in \textbf{occluder} events \cite{baillargeon1991object, spelke1995spatiotemporal} when objects teleported behind one occluder to the back of another disconnected occluder. This event was also modified with different occluder heights in the middle segment \cite{aguiar19992}. Researchers also experimented with inertia violations in \textbf{collision} events \cite{kotovsky1994calibration, kotovsky1998development} where the outcome of the contact was impossible via linear momentum. Finally, gravity violations were also studied as \textbf{support} events \cite{baillargeon1992development, dan2000development, hespos2008young} where imbalanced objects that appeared stable were presented to infants.


We find that Piloto et al. \cite{piloto2018probing}, IntPhys \cite{riochet2018intphys} \& ADEPT \cite{smith2019modeling} to be the most relevant to our work, as they all employ the VoE paradigm in their datasets and evaluation metrics. Additionally, they all present 3D datasets of very similar event categories of physical reasoning.

\textbf{Piloto et al.} \cite{piloto2018probing} presents a 3D VoE dataset of 100,000 training videos and 10,000 pair probes of \textit{surprising} and \textit{expected} videos for evaluation. The dataset categorized their videos into `object persistence', `unchangeableness', `continuity', `solidity' \& `containment'. Their Variational Autoencoder model benchmarked on the dataset and showed promise in `assimilating basic physical concepts'.

\textbf{IntPhys} \cite{riochet2018intphys} is a 3D VoE dataset with 15,000 videos of \textit{possible} events and 3,960 videos of \textit{possible} \& \textit{impossible} events in the test and dev sets. Only three events on `object permanence', `shape constancy' \& `continuity' were present. The study benchmarked the performances of a convolutional autoencoder and generative adversarial network with short \& long-term predictions. The models performed poorly in comparison with their adult human trials but with higher than chance performance.

\textbf{ADEPT} \cite{smith2019modeling} is a model that
uses deep recognition networks, extended probabilistic simulation and particle filtering to predict expectations of objects. They use this model on a 3D VoE dataset of 1,000 training videos of random objects colliding and 1,512 test videos of \textit{surprising} or \textit{control} stimuli. ADEPT accurately predicts the expected location of objects behind an occluder to measure surprise, while replicating adult human judgements on the `how, when \& what' traits of \textit{surprising} scenes \cite{smith2020fine}.

A common finding among developmental psychologists on physical reasoning is that infants create abstract representations of objects \cite{kahneman1992reviewing,gordon1996s} from which they extract spatial features. These high-level features are coupled with high-level rules of reasoning that infants develop over time via process known as explanation-based learning \cite{baillargeon2017explanation} to form their expectation on how a physical scene should play out \cite{lin2020infants}. For instance, the features of a containment event could refer to the heights of the object and container. A rule can be presented as a question: ``is the object taller than the container?'' to which the answer `yes' (based on a simple comparison of the features) would signal an outcome of the object protruding out of the container and vice versa. This guided the construction of AVoE, which provides the ground-truth values for these rules and features in every scene.

\begin{figure*}
    \centering
    \includegraphics[width=\textwidth]{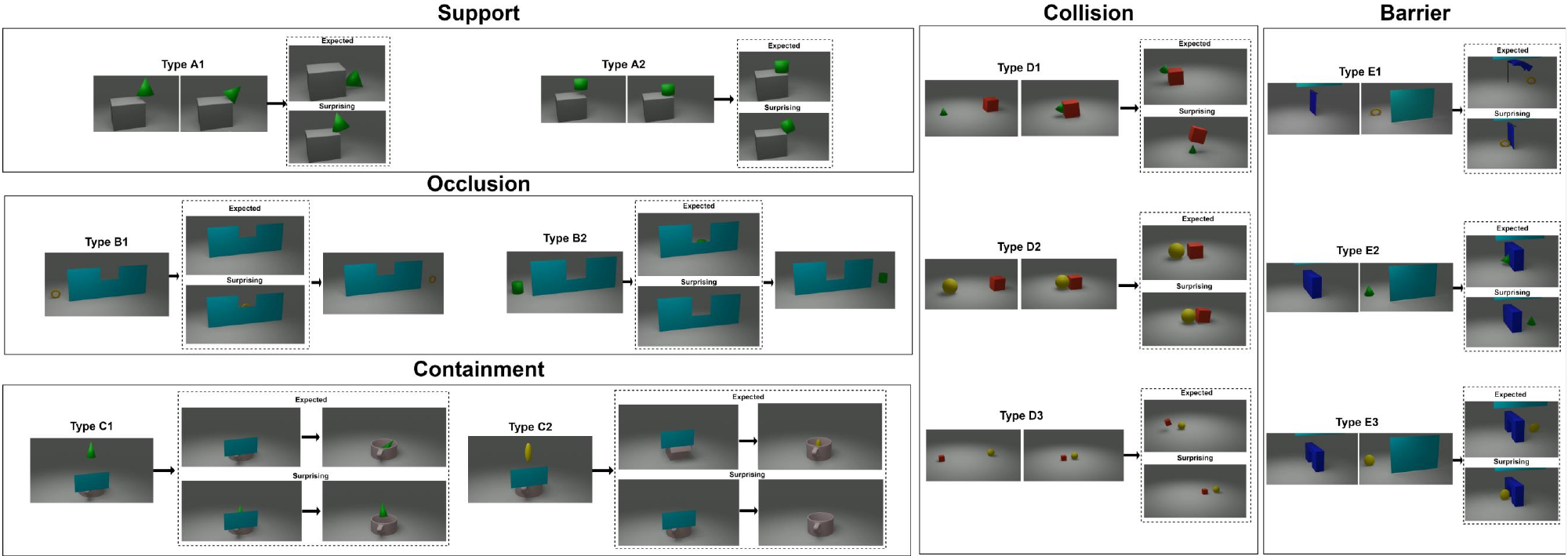}
    \caption{Examples of the different event categories in AVoE with their \textit{surprising} and \textit{expected} outcomes. \textbf{Support}: Type A1 (\emph{originally unbalanced}), Type A2 (\emph{originally balanced}). \textbf{Occlusion}: Type B1 (\emph{object shorter than occluder}), Type B2 (\emph{object taller than occluder}). \textbf{Containment}: Type C1 (\emph{Object fully contained in container}), Type C2 (\emph{object protruding out of container}). \textbf{Collision}: Type D1 (\emph{same speed, different size}), Type D2 (\emph{same speed, same size}), Type D3 (\emph{different speed, same size}). \textbf{Barrier}: Type E1 (\emph{soft barrier}), Type E2 (\emph{solid barrier}), Type E3 (\emph{barrier with opening}).}
    \label{fig:avoe_dataset}
\end{figure*}

\section{AVoE Dataset}

\subsection{Overview}

Figure~\ref{fig:avoe_dataset} comprehensively illustrates the composition of the AVoE dataset, which comprises synthetic video sub-datasets in five event categories: \textbf{support} (A), \textbf{occlusion} (B), \textbf{containment} (C), \textbf{collision} (D) \& \textbf{barrier} (E). Each one of these event categories are split into further sub-categories (e.g. Type B2 etc.) that showcase physical variations based on the differing stimuli of the scene. For each subcategory, every trial showcases an \textit{expected} or \textit{surprising} version of the stimuli. All frames were developed in the open-source 3D graphics software Blender \cite{blender}, using a Python API. Every event category types $\psi \in \{A,B,C,D,E\}$ comes with sets of abstract features $f^\psi$, prior rules $r_{prior}^\psi$ \& posterior rules $r_{post}^\psi$. Prior rules are physical conditions about the event which can be answered with the features (e.g. height, width). These prior rules suffice to answer posterior rules that represent the outcome of the physical interaction. $f^\psi$, $r_{prior}^\psi$ \& $r_{post}^\psi$ are only used for training as VoE evaluation does not test a model's prediction of them. Still, they are still provided for the test and validation sets should researchers wish to evaluate performance in predicting features, prior \& posterior rules.

\textbf{Support (Type A)}: An object is dropped right above the edge of the object. The object's centre of mass is situated either over the edge (\textbf{Type A1}) or within the edge (\textbf{Type A2}).

\textbf{Occlusion (Type B)}: An inert object has an initial momentum to pass behind an occluder. The object can either be shorter than the occluder's middle portion (\textbf{Type B1}) or taller (\textbf{Type B2}).

\textbf{Containment (Type C)}: An inert object falls from a short height above a container, with the interaction hidden behind an occluder. The object is short enough to be fully contained inside the container (\textbf{Type C1}) or tall enough to protude out of the container top (\textbf{Type C2}).

\textbf{Collision (Type D)}: Two inert objects with initial momentum collide head-on. In the first case (\textbf{Type D1}), two objects have the same initial speed with different sizes. In the other two cases, both objects have similar size, with either the different (\textbf{Type D2}) or same (\textbf{Type D3}) initial speeds.

\textbf{Barrier (Type E)}: An inert object has an initial momentum to pass through a barrier, with their interaction hidden behind an occluder. To experiment with different barriers, AVoE comprises events with either a soft barrier (\textbf{Type E1}), a solid barrier (\textbf{Type E2}) or a barrier with opening (\textbf{Type E3}).

\subsection{Procedural Generation}

Multiple physical stimuli that affect the outcome of the interaction were randomly sampled to amplify the diversity of the dataset. Common parameters among the 5 sub-datasets included the object's shape $S_{Obj} \in \{\text{\textit{Cube, Cylinder, Torus, Sphere, Cone, Side Cylinder, Inverted Cone}}\}$ and the object's height \& width, $H_{Obj}, W_{Obj} \in [0.4, 1.6]$, where a $1$ is equivalent to 2$m$ in the 3D environment. The initial contact point of the object on the support in \textbf{A} $C_{Obj} \in [0.2, 0.8]$ where $C_{Obj} = 0.2$ indicates that the 20\% of the object's width is over the edge. In \textbf{B}, the occluder's middle segment height, $H_{Occ}, \in [0.1, 0.9]$ with 1 being the height of the occluder. In \textbf{C}, the container's shape, $S_{Con} \in \{Mug, Box\}$ was also varied with its height and width, $H_{Con}, W_{Con} \in [0.5, 1.5]$. The height ($\propto \text{mass}^{\frac{1}{3}}$) and initial speeds of objects in \textbf{D} were sampled as $H_{Obj} \in [0.5, 1.5]$ \& $V_{Obj} \in [0.5, 2.5]$. In \textbf{E3}, the barrier's opening height and width were sampled in the range, $H_{Bar}, W_{Bar} \in [0.4, 1.4]$.

\subsection{Dataset Structure}

\begin{figure*}
    \centering
    \includegraphics[width=\textwidth]{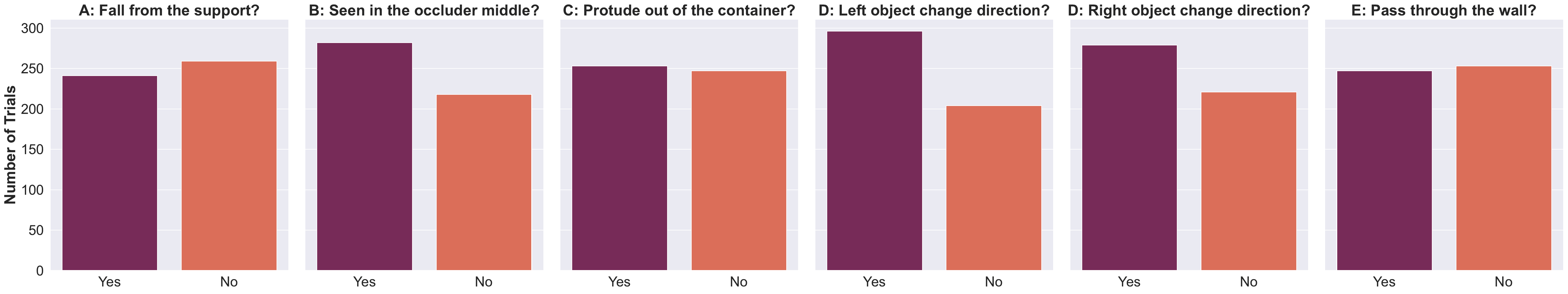}
    \caption{Outcome distribution for AVoE for each one of the event categories (\textbf{D} has 2 outcomes). The distribution depicts `yes' or `no' answers to a question about each outcome for every trial.}
    \label{fig:avoe_outcome}
\end{figure*}

Each event category $\psi$ has 500 different configured trials, amounting to a total of 2,500 configurations in AVoE. 75\% of the trials form the training set which comprises only \textit{expected} videos. 15\% and 10\% of the trials are dedicated to the validation and test sets respectively. The validation and test sets contain both \textit{expected} and \textit{surprising} video pairs with the same stimuli in each trial, keeping the balance at an even 50\%. This sums to 3125 videos. Table~\ref{tab:avoe_count} specifies the raw distribution of AVoE. At 50 frames per video, AVoE offers 156,250 frames, each having a size of $960 \times 540$. Besides the RGB videos, AVoE also provides the depth map and instance segmented frames. Along with the automatically generated ground-truth labels of $f^\psi$, $r_{prior}^\psi$ \& $r_{post}^\psi$ in every video, the frame-wise world position and orientation of all entities are provided. Figure~\ref{fig:avoe_outcome} shows the outcome distribution for each of the 500 stimuli per event category, highlighting that the outcomes are relatively balanced. These outcomes are not forcefully set, rather, they result from the diverse variation of the stimuli.

\begin{table}[h!]
\centering
\caption{The number of trials for each event category in AVoE}
\resizebox{\textwidth}{!}{%
\begin{tabular}{c|ccc|ccc|ccc|cccc|cccc|c}
\hline
\textbf{Set} &
  \multicolumn{3}{c|}{\textbf{Support (A)}} &
  \multicolumn{3}{c|}{\textbf{Occlusion (B)}} &
  \multicolumn{3}{c|}{\textbf{Containment (C)}} &
  \multicolumn{4}{c|}{\textbf{Collision (D)}} &
  \multicolumn{4}{c|}{\textbf{Barrier (E)}} &
  \textbf{Total} \\ \hline
               & A1  & A2  & Total & B1  & B2  & Total & C1  & C2  & Total & D1  & D2 & D3  & Total & E1 & E2 & E3  & Total & -    \\
Train          & 180 & 195 & 375   & 159 & 216 & 375   & 168 & 207 & 375   & 170 & 34 & 171 & 375   & 73 & 63 & 239 & 375   & 1875 \\
Val            & 35  & 40  & 75    & 35  & 40  & 75    & 48  & 27  & 75    & 37  & 4  & 34  & 75    & 14 & 12 & 49  & 75    & 375  \\
Test           & 26  & 24  & 50    & 24  & 26  & 50    & 31  & 19  & 50    & 16  & 9  & 25  & 50    & 8  & 15 & 27  & 50    & 250  \\ \hline
\textbf{Total} & 241 & 259 & 500   & 218 & 282 & 500   & 247 & 253 & 500   & 223 & 47 & 230 & 500   & 95 & 90 & 315 & 500   & 2500 \\ \hline
\end{tabular}%
}
\label{tab:avoe_count}
\end{table}

\section{Conclusion}

In this work, we showcase an alternative approach to predicting VoE in basic physical reasoning events. By leveraging on findings in psychology literature, we propose AVoE, a novel synthetic 3D dataset augmented with ground-truth labels of abstract features and rules for five basic event categories of physical reasoning. These abstract features and rules can be structurally embedded into models as inductive biases to train an artificial agent's ability to discern VoE in physical scenes. AVoE also further splits each event category stimuli into sub-categories for added diversity and complexity. Our recommended evaluation metrics for AVoE are the relative error rate and absolute error rate ($1 - AUC$) as described in \cite{riochet2018intphys}. As future work, we plan to benchmark the performances of baseline generative and probabilistic models along with a novel approach exploiting the feature and rule ground truths of AVoE.

\section*{Acknowledgements}
This research is supported by the Agency for Science, Technology and Research (A*STAR), Singapore under its AME Programmatic Funding Scheme (Award \#A18A2b0046). We would also like to sincerely thank Distinguished Professor Renée L. Baillargeon, Dr Yi Lin and Professor Su-Hua Wang for their consultation and expertise in the domain of infant psychology. Their advice has allowed for the integration of core infant psychology concepts with AI and the creation and improvement of AVoE.

\bibliographystyle{IEEEtran}
\bibliography{references}


\end{document}